\documentclass{article}

\usepackage{natbib}
\setcitestyle{numbers,square}

    \usepackage[preprint]{neurips_2025}

\usepackage[utf8]{inputenc} 
\usepackage[T1]{fontenc}    
\usepackage{hyperref}       
\usepackage{url}            
\usepackage{booktabs}       
\usepackage{amsfonts}       
\usepackage{nicefrac}       
\usepackage{microtype}      
\usepackage{xcolor}         
\usepackage{graphicx}
\usepackage{subfigure}

\definecolor{newcolor}{RGB}{0,163,227}

\title{StyleAR: Customizing Multimodal Autoregressive Model for Style-Aligned Text-to-Image Generation}

\author{
Yi Wu$^1$\thanks{Equal Contribution.}\quad Lingting Zhu$^2$\footnotemark[1]\quad Shengju Qian$^3$\thanks{Project Leader.}\quad Lei Liu$^1$\thanks{Corresponding Authors.}\\ \textbf{Wandi Qiao$^1$\quad Lequan Yu$^2$\footnotemark[3]\quad Bin Li$^1$}
\\
$^1$ University of Science and Technology of China \quad
$^2$ The University of Hong Kong \\
$^3$ The Chinese University of Hong Kong \\
\\
\color{red}{{\href{https://stylear.github.io/}{https://stylear.github.io/}}}
}

\begin{document}

\maketitle

\begin{figure}[!h]
	\centering
	\includegraphics[width=1.0\linewidth]{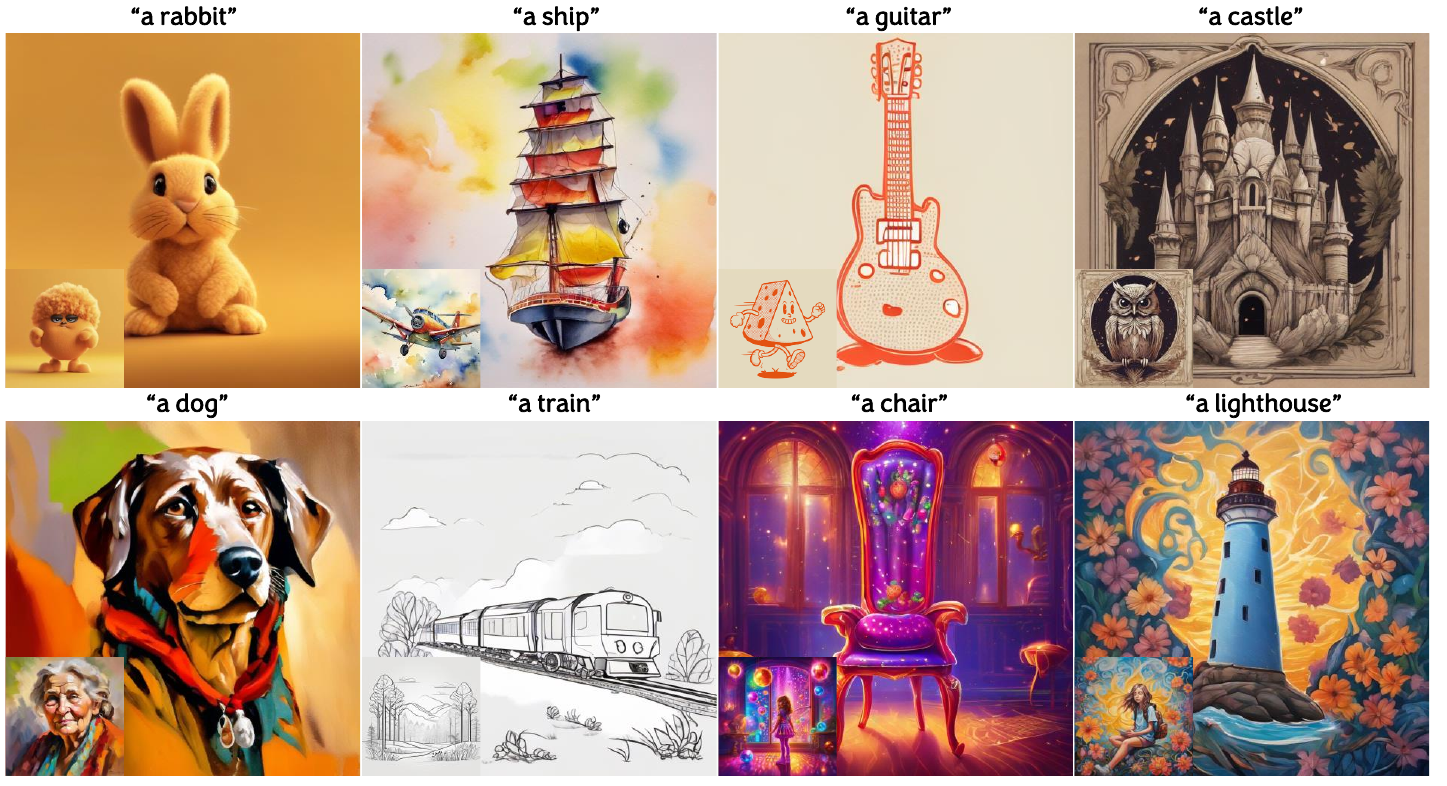}
	\caption{\textbf{Stylized samples of our StyleAR.} Our StyleAR is capable of generating images that are highly consistent in style with the reference images across a diverse range of styles, and highly aligned in semantics with the input prompts of various categories.}
	\label{fig-first}
\end{figure}

\begin{abstract}
In the current research landscape, multimodal autoregressive (AR) models have shown exceptional capabilities across various domains, including visual understanding and generation. However, complex tasks such as style-aligned text-to-image generation present significant challenges, particularly in data acquisition.
In analogy to instruction-following tuning for image editing of AR models, style-aligned generation requires a reference style image and prompt, resulting in a text-image-to-image triplet where the output shares the style and semantics of the input. However, acquiring large volumes of such triplet data with specific styles is considerably more challenging than obtaining conventional text-to-image data used for training generative models.
To address this issue, we propose StyleAR, an innovative approach that combines a specially designed data curation method with our proposed AR models to effectively utilize text-to-image binary data for style-aligned text-to-image generation. Our method synthesizes target stylized data using a reference style image and prompt, but only incorporates the target stylized image as the image modality to create high-quality binary data. To facilitate binary data training, we introduce a CLIP image encoder with a perceiver resampler that translates the image input into style tokens aligned with multimodal tokens in AR models and implement a style-enhanced token technique to prevent content leakage which is a common issue in previous work.
%
Furthermore, we mix raw images drawn from large-scale text-image datasets with stylized images to enhance StyleAR's ability to extract richer stylistic features and ensure style consistency. Extensive qualitative and quantitative experiments demonstrate our superior performance.

\end{abstract}


\section{Introduction}
Multimodal autoregressive (AR) models~\cite{team2024chameleon,liu2024lumina,chern2024anole,wang2024emu3,chen2025janus,sun2024autoregressive} have revolutionized cross-modal content synthesis by using the next-token prediction mechanism, demonstrating superior performance in tasks such as image understanding and text-to-image generation. In particular, in text-to-image generation, AR models outperform traditional diffusion models~\cite{rombach2022high,podell2023sdxl,esser2024scaling,flux2024,xiao2024omnigen,xie2024show,zhou2024transfusion,wu2024infinite} in terms of prompt adherence and image quality. However, when tackling style-aligned text-to-image generation, a task requiring the style and semantic of generated images precise alignment with reference style images and input prompts, respectively, AR models exhibit significant performance gaps compared to diffusion-based approaches~\cite{hertz2024style,wang2024instantstyle,ye2023ip,jeong2024visual,sohn2023styledrop,huang2025artcrafter,liu2023stylecrafter,wang2023styleadapter,ruiz2023dreambooth,gao2024styleshot}. This limitation stems from that the fact that AR models require text-image-to-image triplet data (that is, high-quality annotated data containing reference style images, prompts, and corresponding stylized generated images) in an instruction-following tuning manner to perform style-aligned text-to-image generation, while existing public datasets~\cite{schuhmann2022laion} predominantly provide text-image binary data. 

Existing approaches for style-aligned text-to-image generation which are predominantly diffusion-based and can be broadly categorized into two paradigms. 1) Optimization-free direct inference methods~\cite{ye2023ip,wang2024instantstyle,huang2025artcrafter}, exemplified by IP-Adapter~\cite{ye2023ip} which trains decoupled cross-attention layers to integrate style features with text conditions, allowing zero-shot style transfer. 2) Methods requiring parameter tuning or latent inversion~\cite{jeong2024visual,hertz2024style,ruiz2023dreambooth}, such as Dreambooth~\cite{ruiz2023dreambooth} which performs low-rank~\cite{hu2022lora} adaptation (LoRA) for each new reference style, and StyleAligned~\cite{hertz2024style} which extracts the image feature of reference images via DDIM inversion~\cite{mokady2023null}, then enforces style alignment through shared attention layers. In contrast, AR models remain underexplored for style-aligned text-to-image generation. Compared to diffusion models~\cite{podell2023sdxl,rombach2022high} that flexibly blend multimodal features in latent space, AR models~\cite{liu2024lumina,team2024chameleon,sun2024autoregressive} face dual challenges: their heavy dependence on text-image-to-image triplet data and the inefficiency of extraction of style feature via image tokenizer, which collectively hinder their competitiveness in this domain. 
To overcome these limitations, we present \textbf{StyleAR}, the \textit{first} study to enable AR models to perform style-aligned text-to-image generation. We primary focus on two innovative aspects, i.e., data curation and designs of AR models. 

\begin{figure}[!t]
	\centering
	\includegraphics[width=1.0\linewidth]{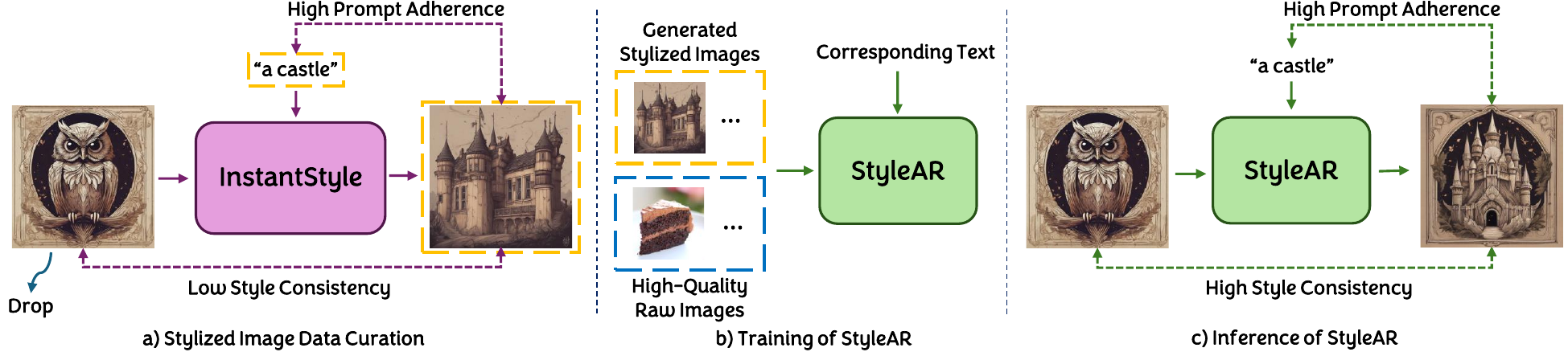}
	\caption{\textbf{The pipeline of our method.} \textbf{a)} We first investigate a novel stylized image data curation to form binary data with high prompt adherence and prevent low style consistency. \textbf{b)} We use a mixed dataset to enhance rich stylistic features learning. \textbf{c)} With the designed data curation and model framework, our method achieve high prompt adherence and style consistency.}
	\label{fig-data}
\end{figure}

On the data aspect, while we can use diffusion-based methods for creating triplet data for instruction-following tuning, they suffer from quality issues due to two main reasons. First, the generated stylized images exhibit low style consistency with the input reference style image, which limiting dataset quality. Second, the generated triplet data serve as ground truth for instruction-following tuning, making the capability frontier of diffusion models an upper bound for AR models, thus challenging the purpose of tailoring AR models for style-aligned text-to-image generation. To address these issues, we propose to drop the reference style image in the data generation process and use only the prompt and the generated stylized image for binary data construction, illustrated as Fig.~\ref{fig-data}(a). Moreover, we discover that mixing raw images drawn from large-scale text-image dataset~\cite{schuhmann2022laion} with stylized images at an optimal 3:1 ratio during training stage can significantly improve our StyleAR's ability to extract richer stylistic features, illustrated as Fig.~\ref{fig-data}(b)(c). Using these curated stylized image data along with raw image data, we can create a high-quality dataset for AR model tuning.

The remaining part focuses on customizing AR models for style-aligned image generation using binary data. We introduce two key innovations to enhance the AR model to learn from binary data and improve inference performance. \textbf{1)} We integrate the image encoder~\cite{radford2021learning} with a perceiver~\cite{jaegle2021perceiver,alayrac2022flamingo} resampler module to convert the input image to a unified token space that integrates effectively with multimodal tokens. \textbf{2)} We introduce a style-enhancing mechanism for tokens using SAM~\cite{kirillov2023segment} and Gaussian noise injection to remove irrelevant semantic features from the reference style image.

The key contributions of this work can be summarized as follows:
\begin{itemize}
\setlength{\itemsep}{2pt}
\setlength{\parsep}{2pt}
\setlength{\parskip}{2pt}
    \item We investigate a novel data curation method for crafting binary stylized text-image data from a diffusion counterpart model for training AR models, achieving high quality and avoiding upper bound limit of the base data generation model.
    
    
    \item We propose a training framework for AR models to use binary text-to-image data to perform stylized-aligned text-to-image generation, preventing the difficulty of scaling up text-image-to-image triplet data for instruction-following tuning.
    
    \item We also propose the style-enhanced tokens technique, which can effectively solve the problem of content leakage in the stylized-aligned text-to-image generation task and significantly improve prompt adherence and style consistency.

    \item Extensive quantitative evaluation, qualitative experiments, and user study demonstrate that our StyleAR achieves state-of-the-art performance in both prompt adherence and style consistency, surpassing existing diffusion-based approaches. Moreover, StyleAR integrates additional conditions effectively, such as depth map and other structural control.

\end{itemize}

\section{Related Work}
\subsection{Multimodal Understanding and Generation Unified Models}
In recent times, unified models~\cite{team2024chameleon,liu2024lumina,chern2024anole,wang2024emu3,chen2025janus,zhou2024transfusion,tang2024hart,fan2024fluid,xie2024show,xiao2024omnigen,wu2025proxy,sun2025personalized} for multimodal generation and understanding have achieved significant development. These works fall primarily into two categories: autoregressive-based models represented by Chameleon~\cite{team2024chameleon}, and diffusion-based models exemplified by Transfusion~\cite{zhou2024transfusion}. In the first category, Chameleon~\cite{team2024chameleon} pioneered an early fusion approach in which all multimodal inputs are projected onto a shared representational token space from the start, training a transformer to process sequences of multimodal tokens. Building on this foundation, Lumina-mGPT~\cite{liu2024lumina} enhances training datasets and achieves flexible image generation of varying aspect ratios, and Emu3~\cite{wang2024emu3} adopts a two-stage training framework and introduces the DPO~\cite{rafailov2023direct} (Direct Preference Optimization) algorithm to improve visual quality and prompt alignment. Furthermore, Janus-Pro~\cite{chen2025janus} decouples visual encoding for image understanding and image generation tasks, employing different image encoders for each objective. HART~\cite{tang2024hart} and FLUID~\cite{fan2024fluid} explore integrating continuous tokens into AR models to improve the quality of image generation. In the second category, Transfusion~\cite{zhou2024transfusion} innovatively combines autoregressive and diffusion training paradigms within a single transformer framework: images are trained via diffusion process, while text follows next-token prediction pattern. Furthermore, Show-o~\cite{xie2024show} and OmniGen~\cite{xiao2024omnigen} expand this approach by scaling the training datasets. In particular, OmniGen~\cite{xiao2024omnigen} incorporates diverse computer vision tasks (e.g., inpainting, deblurring) and subject-driven image generation task into its training data, enabling strong performance across multiple common benchmarks.

\subsection{Style-Aligned Text-to-Image Generation}
Existing approaches for style-aligned text-to-image generation are predominantly diffusion-based and can be broadly categorized into two paradigms. 1) Optimization-free methods~\cite{ye2023ip,wang2024instantstyle,huang2025artcrafter}, exemplified by IP-Adapter~\cite{ye2023ip} which trains decoupled cross-attention layers to dynamically integrate style features from pretrained image encoders~\cite{radford2021learning} with text conditions, allowing zero-shot style transfer. Similarly, InstantStyle~\cite{wang2024instantstyle} identifies content-specific and style-specific layers in Stable Diffusion XL~\cite{podell2023sdxl} model, proposing a layer-wise decoupling mechanism to suppress content leakage from the reference style images. 2) Methods~\cite{jeong2024visual,hertz2024style,ruiz2023dreambooth} require parameter tuning or latent inversion~\cite{mokady2023null}, such as 
StyleAligned~\cite{hertz2024style} which extracts image features via DDIM inversion~\cite{mokady2023null} and enforces stylization through attention layers. In contrast, AR models remain underexplored for style-aligned text-to-image generation. Compared to diffusion models~\cite{podell2023sdxl,rombach2022high} that flexibly blend multimodal features in latent space, AR models~\cite{liu2024lumina,team2024chameleon,sun2024autoregressive} face dual challenges: heavy dependence on instruction-following data and inefficiency of style feature extraction in next-token prediction frameworks via image tokenizer, which hinder their competitiveness.

\section{Method}

\begin{figure}[!t]
    \centering
    \includegraphics[width=0.95\linewidth]{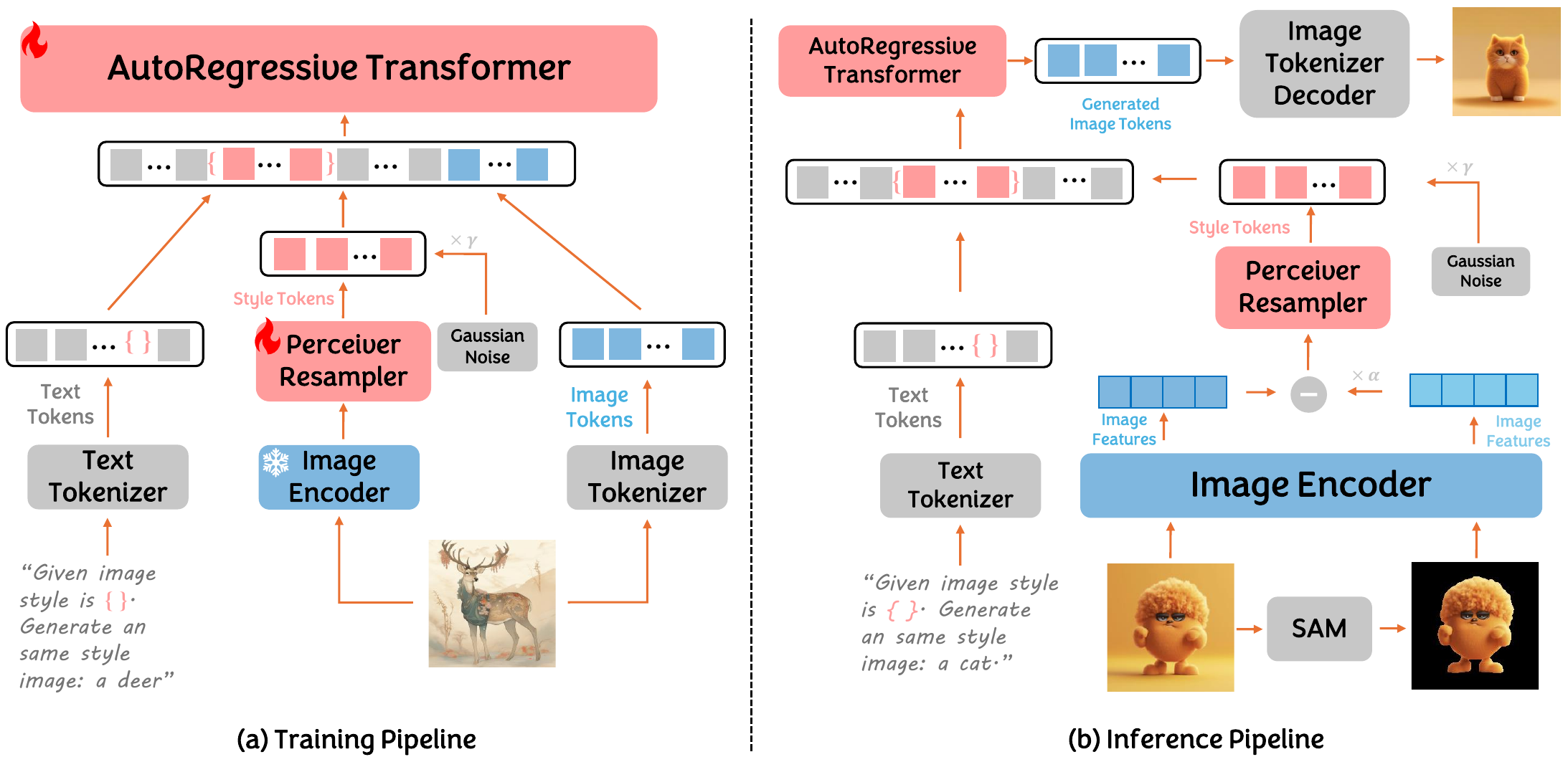}
    \caption{\textbf{The framework of our StyleAR.} During training, we utilize a frozen CLIP~\cite{radford2021learning} image encoder along with a trainable perceiver~\cite{jaegle2021perceiver,alayrac2022flamingo} resampler module to efficiently extracted features. Subsequently, style tokens are combined with the injected Gaussian noise and concatenated with multimodal tokens by replacing the placeholder tokens. During inference, we incorporate SAM~\cite{kirillov2023segment} to remove irrelevant semantic contents in the reference style image.}
    \label{fig-framework}
\end{figure}

\subsection{Preliminaries}
During the image generation training process of the AR model, for an input image $x\in\mathbb{R}^{H\times W\times 3}$, it is first quantized into $q\in\mathbb{Q}^{h\times w}$ discrete tokens by an image tokenizer~\cite{van2017neural,esser2021taming,yu2021vector}, where $h=H/p_h$, $w=W/p_w$, $p_h$ and $p_w$ are the downsample ratios of the image tokenizer in the vertical and horizontal directions respectively, and $q^{(i,j)}$ is the indices of the image codebook. Then the image tokens are flattened into a sequence of length $h\times w$ and is concatenated with the text tokens passed into a Transformer-based~\cite{vaswani2017attention} autoregressive model for training. During the inference stage, given the text tokens $t$, the autoregressive model can generate image tokens through next-token prediction:
\begin{equation}
    \Pi^{h\cdot w}_{t=1}p(q_t|q_{<t},t).
\end{equation}
Finally, the generated image tokens are converted to pixel space via a image decoder.

\subsection{Data Curation}
A primary contribution of our method focuses on the data curation part. If we aim at creating triplet data for instruction-following tuning, while we can use InstantStyle~\cite{wang2024instantstyle} to create such data, they suffer from low style consistency and make the capability frontier of diffusion models an upper bound for AR models. In contrast, we drop the reference style image in the data generation process and use only the prompt and the generated stylized image for binary data construction. In this way, we acquire high-quality stylized binary data and prevent low style consistency. Moreover, through practical experiments, we have found that if we rely solely on this stylized dataset for model training, the model's ability to capture image features during the inference stage is unsatisfactory (as shown in Section~\ref{sec-ablation}), which leads to the style consistency staying low between the generated images and the reference style images. Moreover, considering model nature during pre-training phase of the text-to-image generation task, the training sets it uses mostly consist of raw images that are not stylized. If we only use the stylized dataset for training, the domain gap brings difficulties to the model training process. In view of this, when training our StyleAR, we simultaneously use the raw image dataset and the stylized image dataset in a certain proportion to serve as training dataset. 

\subsection{Framework of StyleAR}

\noindent \textbf{Training with Binary Data.} 
The framework of our StyleAR are shown in Figure~\ref{fig-framework}. 
To enable binary data training, we design the model to use the input image in a self-supervised manner, extracting style features and learning to predict the image tokens of the same image. In particular, the input image $I$ is first processed through CLIP~\cite{radford2021learning} image encoder $E_I$ to extract image features. The image features are converted to style tokens $s\in \mathbb{R}^{M\times C}$, which fit the unified token space of the AR model by a perceiver~\cite{jaegle2021perceiver,alayrac2022flamingo} resampler module $R$, where $M=16$ is the number of style tokens and $C=4,096$ is the dimension of the unified token space of the AR model. Additionally, to alleviate the content leakage issue~\cite{jeong2024visual}, we inject Gaussian noise $n$ to style tokens to weaken irrelevant semantic features and enforce the AR model to pay attention to the semantic information from prompts during image generation. The process of image tokens generation is formulated as $\Pi^{h\cdot w}_{t=1}p(q_t|q_{<t},t,\hat{s})$, where $\hat{s}=s+\gamma\cdot n$, and $\gamma$ is the strength of Gaussian noise injection.

\noindent \textbf{Style-Enhanced Inference.}
We incorporate the SAM (Segment Anything Model)~\cite{kirillov2023segment}, represented as $S_I$, combining with the Gaussian noise injection mechanism to form the \textit{style-enhanced tokens} technique to further reduce the risk of content leakage and facilitate accurate and reliable inference. Concretely, the input image $I$ and its segmented image $I_S$ are passed through the CLIP~\cite{radford2021learning} image encoder to obtain the corresponding image features $F$ and $F_S$. Via feature subtraction $F-F_S$ to exclude semantic information, the result is mapped to the unified token space through the perceiver resampler module. Moreover, to preserve more fine-grained style features, we also introduced a residual path in the unified token space. The style-enhanced tokens $\hat{s}_e$ can be formulated as follows:
\begin{equation}
    \hat{s}_e = \alpha\cdot R(F) + (1-\alpha)\cdot R(F-F_S) + \gamma\cdot n,
\end{equation}
where $\alpha$ is the residual ratio of the residual path. This inference mechanism significantly improves the stylized image quality, achieving high prompt adherence and high style consistency.


\noindent \textbf{Post-Training.}
Recent numerous research~\cite{ouyang2022training,rafailov2023direct,shao2024deepseekmath} have demonstrated the potential of post-training, mainly via reinforcement learning, to enhance the reasoning capabilities of large language models (LLMs) and human preference alignment. In the field of image generation, whether in diffusion models~\cite{black2023training,fan2023dpok,miao2024subject} or AR models~\cite{wang2024emu3,wang2025simplear}, post-training is frequently used to improve prompt alignment and visual quality of generated images~\cite{xu2023imagereward, xue2025dancegrpo, wallace2024diffusion, fan2023dpok}. In this work, we utilize Direct Preference Optimization~\cite{rafailov2023direct} (DPO) algorithm to boost the prompt alignment in the style-aligned text-to-image generation. Specifically, we implement a standard DPO strategy via ranking data creation. For each prompt, we use our StyleAR to generate two images and use the VLM~\cite{qwen2,qwen2.5} to select the image that better aligns with the semantics of the corresponding prompt from these two images. Based on the scoring results, we construct a triplet $(p_i,x_i^{chosen},x_i^{rejected})$ for DPO training.

\section{Experiments}
\subsection{Experimental Details}
\textbf{Details of Model and Dataset.} Our StyleAR is implemented based on the FP-SFT@768 version of Lumina-mGPT~\cite{liu2024lumina}. The raw image training dataset is sourced from the open source dataset~\cite{opendiffusionai}. For the stylized image training dataset, we collect 80 distinct artistic style images from the open source dataset~\cite{xiao2024omnigen,wikiart} and generate 200 semantically diverse images per style using InstantStyle~\cite{wang2024instantstyle}, resulting in a total training dataset of 16,000 stylized image data. During each epoch, we randomly sample 10\% (49,368 images) data from the raw image dataset and mix with the full stylized image dataset to construct the training dataset. The training configuration employs a batch size of 64 with a learning rate of 2e-5 and the rank of LoRA~\cite{hu2022lora} parameters employed in the AR model is set to 4.

\textbf{Details about Evaluation Metrics.} Following previous work~\cite{hertz2024style,liu2023stylecrafter,sohn2023styledrop,huang2025artcrafter}, we use the CLIP-T~\cite{radford2021learning} metric to evaluate prompt adherence, which is the cosine similarity between the CLIP text embeddings of the input prompts and the CLIP image embeddings of the corresponding generated images. CLIP-I and DINO~\cite{zhang2022dino} metric are used to evaluate style consistency, which is the cosine similarity between the image embeddings of the reference style images and the image embeddings of the corresponding generated images. To robustly measure performance and generalization capabilities of the methods, we collected 10 diverse reference style images and 20 various prompts including human activities, animals, buildings, vehicle, musical instruments, and furniture. For the evaluation suite, we generate four images per style and per prompt, totaling 800 images.

\begin{figure}[!t]
	\centering
	\includegraphics[width=0.92\linewidth]{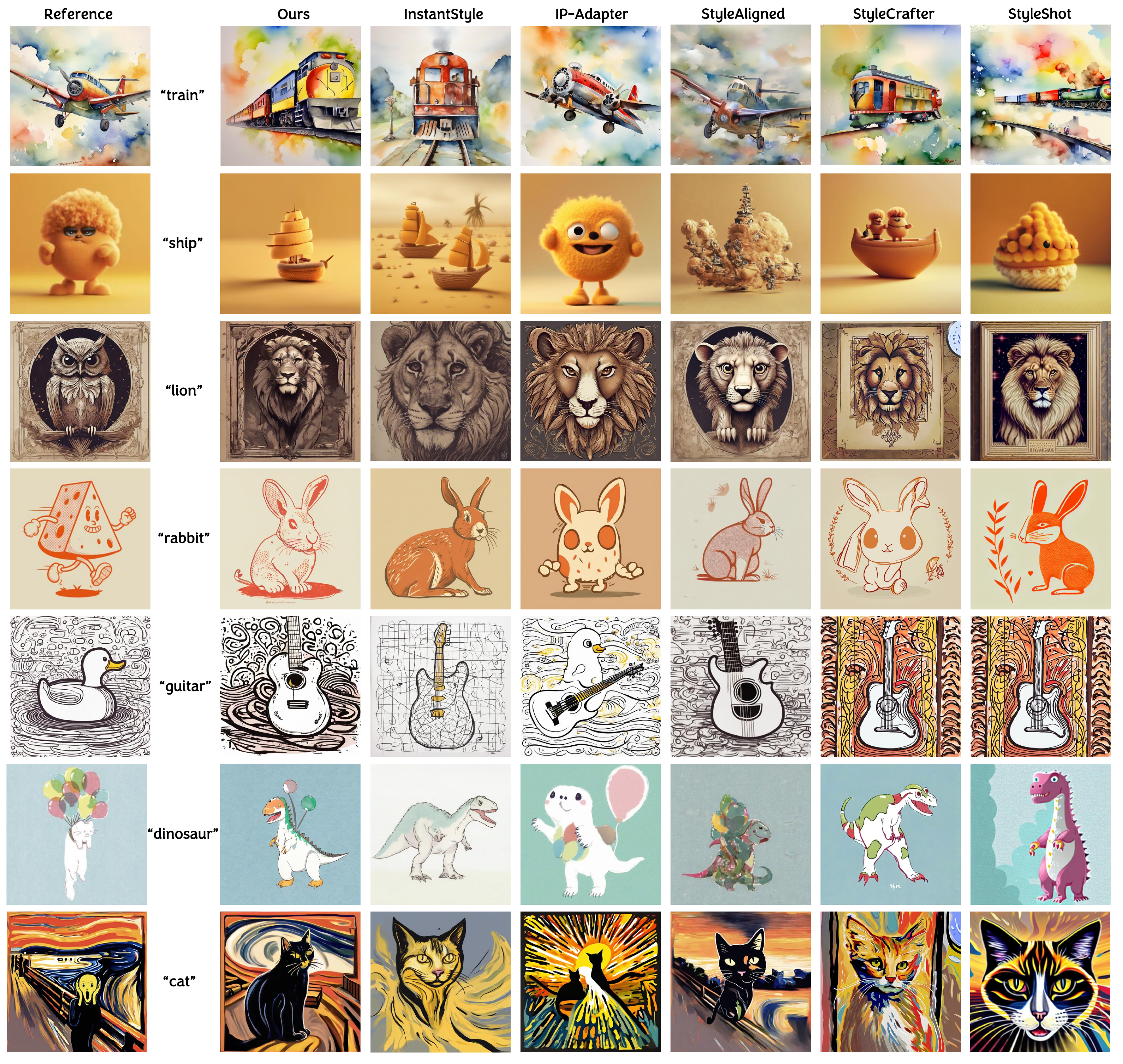}
	\caption{\textbf{Qualitative comparison.} We conducted a comprehensive qualitative evaluation by comparing our StyleAR with various existing methods which are all diffusion-based, including InstantStyle~\cite{wang2024instantstyle}, IP-Adapter~\cite{ye2023ip}, StyleAligned~\cite{hertz2024style}, StyleCrafter~\cite{liu2023stylecrafter}, StyleShot~\cite{gao2024styleshot}.}
	\label{fig-results}
\end{figure}

\subsection{Comparisons}
We perform a comprehensive comparison between our proposed StyleAR (AR-based) with existing methods which are diffusion-based, including InstantStyle~\cite{wang2024instantstyle}, IP-Adapter~\cite{ye2023ip}, StyleAligned~\cite{hertz2024style}, StyleCrafter~\cite{liu2023stylecrafter}, and StyleShot~\cite{gao2024styleshot}. With the exception of StyleShot~\cite{gao2024styleshot}, which uses the implementation of the Stable Diffusion 1.5~\cite{rombach2022high} implementation, all other comparative methods are based on the Stable Diffusion XL architecture~\cite{podell2023sdxl}. All comparative experiments were conducted using the official open source implementations of the baseline methods, with hyperparameter settings strictly adhering to the configurations prescribed in their respective technical documentations.

\textbf{Qualitative Comparison.} The quantitative comparison is demonstrated in Figure~\ref{fig-results}. According to the results, InstantStyle~\cite{wang2024instantstyle} shows a superior prompt adherence, achieving a notable semantic alignment of the input prompt and the generated image. However, it shows an inferior style consistency between the generated images and the reference style images. The IP-Adapter~\cite{ye2023ip} often exhibits a failure in prompt adherence, where the generated images deviate from the input prompts. As demonstrated in the first row of Figure~\ref{fig-results}, given the prompt ``a train'', the model incorrectly generates an aircraft image. This artifact originates from content leakage, a phenomenon in which the semantic content of the reference style image (e.g., aircraft) inappropriately propagates into the image generation process via decoupled cross-attention layers in the IP-Adapter~\cite{ye2023ip}, overriding the semantic content of the input prompts. StyleAligned~\cite{hertz2024style} demonstrates unstable generation outcomes and semantic chaos due to the inherent limitations of its shared attention layers in achieving stable decoupling between content and style features. StyleCrafter~\cite{liu2023stylecrafter} and StyleShot~\cite{gao2024styleshot} demonstrate satisfactory prompt adherence, but exhibit notable deficiencies in style consistency. In contrast, our StyleAR demonstrates excellent prompt adherence and accurately captures the overall and detailed features of the reference style.

\textbf{Quantitative Comparison.} The quantitative comparison results are shown in Table~\ref{tab-results}. Our StyleAR achieves a superior balance between prompt adherence and style consistency. On the one hand, StyleAR attains the second highest performance in prompt adherence, slightly below InstantStyle~\cite{wang2024instantstyle}, while InstantStyle shows poor style consistency. On the other hand, StyleAR ranks second in CLIP-I and DINO metrics, marginally behind IP-Adapter~\cite{ye2023ip}. However, IP-Adapter~\cite{ye2023ip} suffers from poor prompt adherence and severe content leakage (as shown in Figure~\ref{fig-results} of qualitative results), which lead to abnormal increases in CLIP-I and DINO metrics. In contrast, our method effectively extracts the style features of the reference style image and generate the target images without any content leakage.

\begin{table}[!t]
  \caption{\textbf{Quantitative comparison.} We conduct a comprehensive qualitative evaluation by comparing our StyleAR with various existing methods, including InstantStyle~\cite{wang2024instantstyle}, IP-Adapter~\cite{ye2023ip}, StyleAligned~\cite{hertz2024style}, StyleCrafter~\cite{liu2023stylecrafter}, StyleShot~\cite{gao2024styleshot}. The CLIP-T metric reflects the prompt adherence. The CLIP-I and DINO metrics indicate the style consistency. Notably, the content leakage issue of IP-Adapter leads to a abnormal increase in CLIP-I and DINO metrics.}
  \label{tab-results}
  \centering
  \resizebox{1.0\textwidth}{!}{
  \begin{tabular}{ccccccc}
  \hline
         & StyleAR (Ours)   & InstantStyle  &StyleAligned&StyleCrafter  &IP-Adapter &StyleShot\\
\hline
    CLIP-T ($\uparrow$)   & \underline{0.2893}    & \textbf{0.2944}&0.2778&0.2815&0.2655&0.2807     \\
    CLIP-I ($\uparrow$)   &\underline{0.7456}    & 0.6899&0.7003&0.7098&\textbf{0.7971}&0.6888      \\
    DINO ($\uparrow$)     & \underline{0.6136}  &0.4902&0.6080&0.6094   & \textbf{0.6617}&0.5668  \\
    \hline
  \end{tabular}}
\end{table}


\begin{figure}[!t]
	\centering
	\includegraphics[width=1.0\linewidth]{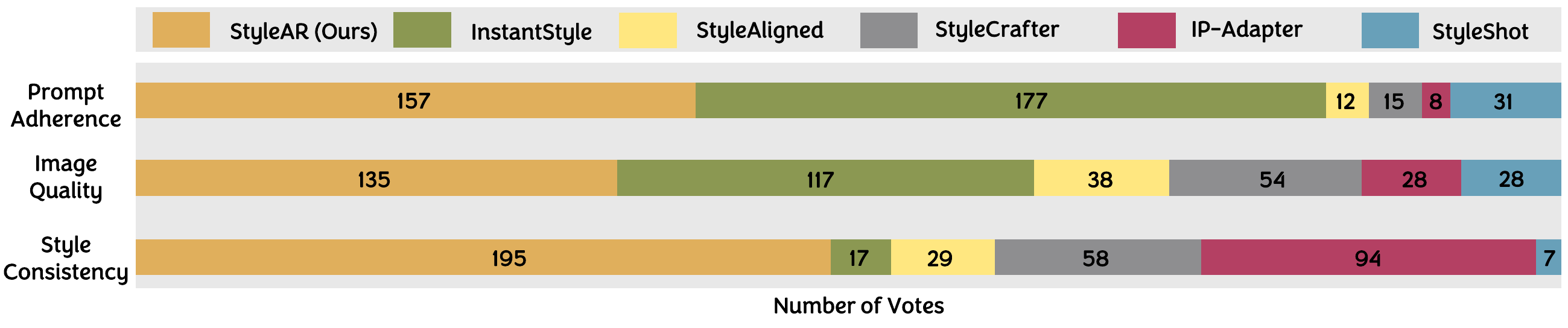}
	\caption{\textbf{User study.} We conducted a user study by comparing StyleAR with existing methods, including InstantStyle~\cite{wang2024instantstyle}, IP-Adapter~\cite{ye2023ip}, StyleAligned~\cite{hertz2024style}, StyleCrafter~\cite{liu2023stylecrafter}, StyleShot~\cite{gao2024styleshot}.}
	\label{fig-user}
\end{figure}

\textbf{User Study.} The results of the user study are shown in Figure~\ref{fig-user}. In terms of prompt adherence and image quality, our method performs on par with InstantStyle~\cite{wang2024instantstyle}, both significantly outperforming other methods. Moreover, in terms of style consistency, our method far surpasses all others. In contrast, the InstantStyle~\cite{wang2024instantstyle} method exhibits poor style consistency. It can be seen that our method not only strictly adheres to the input prompts to generate high-quality images but also ensures a high degree of style consistency between the generated images and the reference style images.

\begin{figure}[!t]
	\centering
	\includegraphics[width=0.95\linewidth]{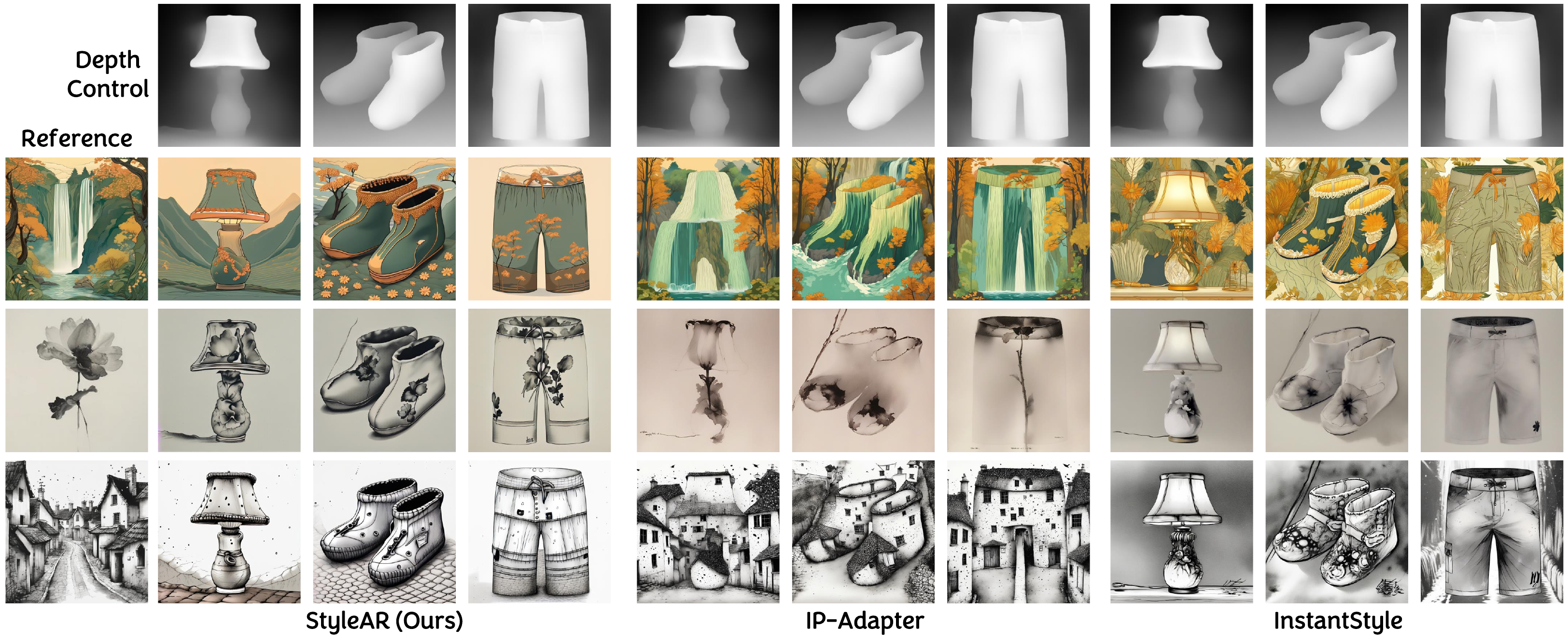}
	\caption{\textbf{Qualitative comparison of integration with additional conditions.} We show the comparison results of control generation with Ours and diffusion models.}
	\label{fig-controlnet}
\end{figure}

\textbf{Additional Results.} Freezing the original parameters of the AR model, StyleAR can retain the original generation capabilities of the AR model, such as structural control. Compared with diffusion-based methods like IP-Adapter~\cite{ye2023ip} and InstantStyle~\cite{wang2024instantstyle}, our StyleAR method performs better on image quality, condition fidelity, and style consistency, shown in Figure~\ref{fig-controlnet}.

\begin{figure}[!t]
	\centering
	\includegraphics[width=0.8\linewidth]{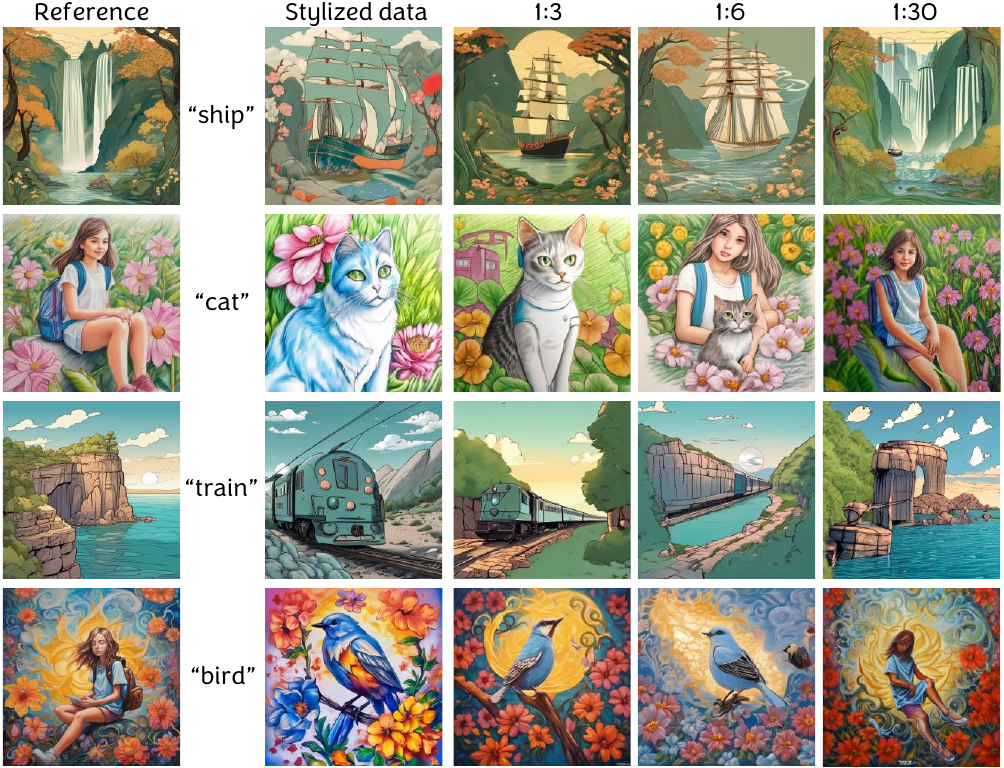}
	\caption{\textbf{Ablation study of composition of training datasets.} We investigate the impact of the composition of training datasets in our StyleAR. The compared training datasets include pure stylized image data, the ratios of stylized image data to raw image data are 1:3, 1:6 and 1:30 respectively.}
	\label{fig-abla-datasets}
\end{figure}

\subsection{Ablation Study}
\label{sec-ablation}
In this section, we conduct ablation experiments to examine how the training dataset's elements and the design modules impact StyleAR's results.

\noindent \textbf{Composition of training dataset.} We have meticulously designed different compositions of the training datasets to conduct ablation experiments, in order to explore the impact of the training datasets on StyleAR. Specifically, the compared training datasets include pure stylized image data, the ratios of stylized image data to raw image data are 1:3, 1:6 and 1:30 respectively. The qualitative results are shown in Figure~\ref{fig-abla-datasets} and the quantitative results are shown in Table~\ref{tab-abla-datasets}. According to the results, when the training dataset only contains stylized image data, prompt adherence is relatively good, but style consistency is relatively poor. In contrast, when raw images are added, specifically when the ratio of stylized image data to raw image data is 1:3, the style consistency is significantly improved and the prompt adherence also remains at a good level. However, when the ratios are further increased to 1:6 and 1:30, the content leakage occurred and the generated images show ``overfitting'' to the reference style images. The irrelevant semantic content also appears in the generated images, causing the semantics of the generated images to not follow the semantics of the input prompt. Consequently, we conclude that in stylization tasks, in addition to stylized image data, appropriately adding some raw image data can improve style consistency and will not degrade the prompt adherence.

\begin{figure}[!t]
	\centering
	\includegraphics[width=0.88\linewidth]{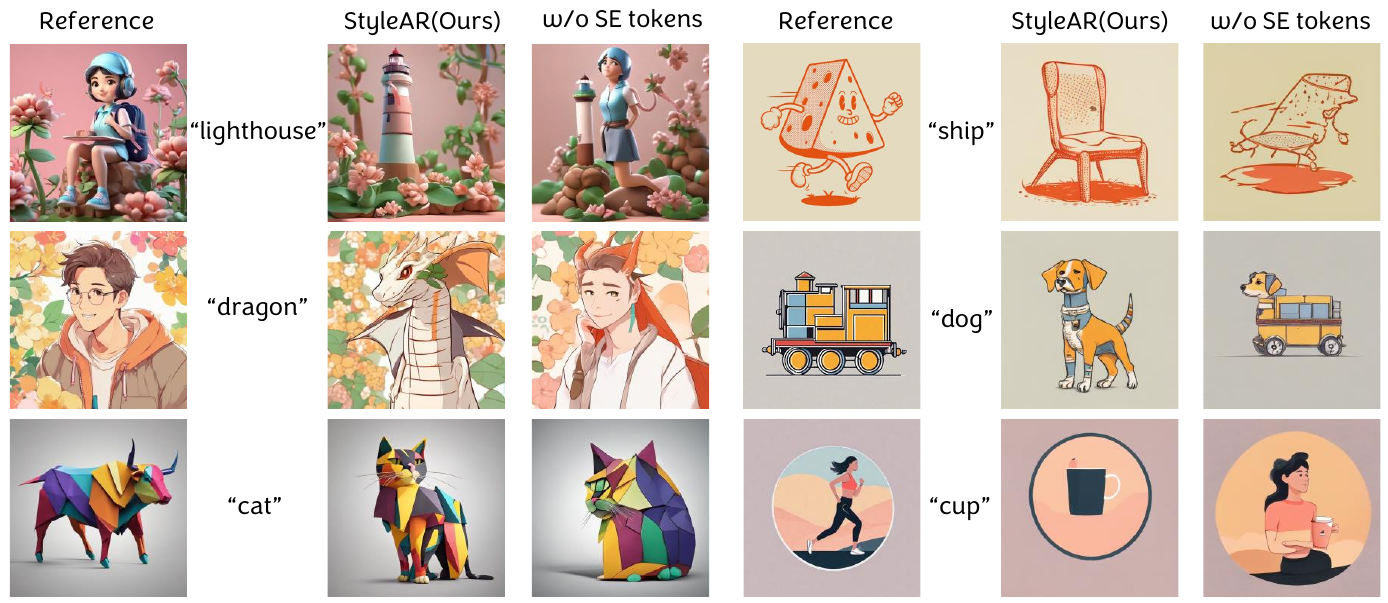}
	\caption{\textbf{Ablation study of style-enhanced (SE) tokens.} We investigate the style-enhanced tokens technique in our StyleAR. The ablation study demonstrates the effectiveness.}
	\label{fig-abla-se}
\end{figure}

\begin{table}[!t]
  \caption{\textbf{Quantitative ablation study of composition of training dataset, style-enhanced (SE) tokens and DPO post-training.} The compared variants covers the training dataset study and module designs. The ranking pattern is similar as that in Table~\ref{tab-results}, indicating the content leakage phenomenon.}
  \label{tab-abla-datasets}
  \centering
  \resizebox{1.0\textwidth}{!}{
  \begin{tabular}{ccccccc}
  \hline
         & Stylized data   &Ratio 1:3 (Ours)  & Ratio 1:6& Ratio 1:30 & Ours w/o SE tokens & Ours w/o DPO \\
\hline
    CLIP-T ($\uparrow$)   &  \textbf{0.2970}   &\underline{0.2893} &0.2688&0.2170 &0.2683  &0.2874  \\
    CLIP-I ($\uparrow$)   &  0.7022  &0.7456 &0.7475&\textbf{0.7798}  & \underline{0.7552} &0.7151  \\
    DINO ($\uparrow$)     & 0.5241  &0.6136&0.6657&\textbf{0.7677}   &\underline{0.6765}&0.6193 \\
    \hline
  \end{tabular}}
\end{table}

\textbf{Impact of Style-Enhanced Tokens.} To evaluate the effectiveness of our proposed style-enhanced tokens technique, we conduct quantitative and qualitative ablation study to compare our StyleAR and our StyleAR without style-enhanced tokens. The results are shown in Figure~\ref{fig-abla-se} and in the six columns of Table~\ref{tab-abla-datasets}. According to the results, when without style-enhanced tokens, the irrelevant semantic features from reference style images show in the generated images, which leads to the generated images not conforming to the semantic control of the input prompt and the situation of chaotic generation occurs. In contrast, prompt adherence and image quality are both improved when style-enhancing mechanism is employed, which enables the style-enhanced tokens to assist the model in filtering out the irrelevant semantic information of reference style images, ensuring that generated images are highly consistent with the input prompt and significantly improving image quality.

\textbf{Impact of DPO.} To evaluate the effectiveness of DPO post-training in our StyleAR, we conduct quantitative ablation study to compare our StyleAR and our StyleAR without DPO pose-training. The quantitative results are shown in the seventh column of Table~\ref{tab-abla-datasets}. DPO post-training can improve the prompt adherence and slightly enhance the style consistency of StyleAR.

%

\section{Conclusion and Limitations}
In this work, we present StyleAR, the first work to use image-text binary data to enable the multimodal autoregressive model to perform style-aligned text-to-image generation which is dominated by diffusion-based methods. Compared with using triplet data in instruction-following tuning of previous AR models, our use of image-text binary data training can easily scale up the size of the training dataset, thereby improving the model's performance. Furthermore, ablation experiments verify the effectiveness of our module designs, including stylized-raw mixed training strategy and style-enhanced tokens technique to enhance style consistency and prompt adherence. However, the current implementation requires depth map extraction for content control rather than direct input of content image to realize style transfer. Future research will focus on leveraging the multimodal input capabilities of autoregressive models to enable simultaneous integration of style reference images and content-specific visual image inputs for further conditional image generation.

\section{Acknowledgments}
This research is supported by the Anhui Provincial Natural Science Foundation (Grant No.2408085QF214), the Fundamental Research Funds for the Central Universities (Grant No.WK2100000045), the Opening Project of the State Key Laboratory of General Artificial Intelligence (Grant No.SKLAGI2024OP10, Grant No.SKLAGI2024OP11), the Research Grants Council of Hong Kong (C5055-24G and T45-401/22-N) and in part by the National Natural Science Foundation of China (No. 62201483).

\newpage
\small
\bibliographystyle{ieee_fullname}
\bibliography{neurips_2025}

\end{document}